\documentclass[graybox]{svmult}

\usepackage{type1cm}        
\usepackage{makeidx}         
\usepackage{graphicx}        
\usepackage{multicol}        
\usepackage[bottom]{footmisc}

\usepackage{newtxtext}       %
\usepackage{newtxmath}       
\usepackage{amssymb}

\usepackage{xspace}
\usepackage{graphicx}
\usepackage{paralist}
\usepackage{url}
\usepackage{subfig}

\usepackage{algorithm}
\usepackage{algorithmicx}
\usepackage{algpseudocode}

\usepackage{xcolor}
\usepackage{colortbl}
\usepackage{amssymb}
\definecolor{GAINSBORO}{HTML}{DCDCDC}
\definecolor{LIGHTYELLOW}{HTML}{FFFFE0}
\definecolor{MISTYROSE}{HTML}{FFE4E1}
\definecolor{ALICEBLUE}{HTML}{F0F8FF}
\usepackage{pifont}
\algblock{ParFor}{EndParFor}
\algnewcommand\algorithmicparfor{\textbf{parfor}}
\algnewcommand\algorithmicpardo{\textbf{do}}
\algnewcommand\algorithmicendparfor{\textbf{end\ parfor}}
\algrenewtext{ParFor}[1]{\algorithmicparfor\ #1\ \algorithmicpardo}
\algrenewtext{EndParFor}{\algorithmicendparfor}

\algrenewcommand\alglinenumber[1]{\tiny #1:}

\newcommand{\method}{\texttt{Redux-Lipizzaner}\xspace}

\newcommand{\horse}{\texttt{Lipizzaner}\xspace}

\newcommand{\xGrid}[1]{#1$\times$#1}

\newcommand{\nameAlgCell}{\texttt{CoevolveAndTrainModels}\xspace}
\newcommand{\nameAlgLipiTraining}{\texttt{LipizzanerTraining}\xspace}

\newcommand{\nameAlgMixture}{\texttt{MixtureEA}\xspace}
\newcommand{\alg}{\texttt{Algorithm}\xspace}

\renewcommand{\v}[1]{\mathbf{#1}}

\DeclareMathOperator*{\argmin}{argmin}

\makeindex             

\begin{document}

\title*{Data Dieting in GAN Training}
\author{Jamal Toutouh, Erik Hemberg, and Una-May O'Reilly}
\institute{Jamal Toutouh \at MIT CSAIL, Cambridge, MA, USA \email{toutouh@mit.edu}
\and Erik Hemberg \at MIT CSAIL, Cambridge, MA, USA \email{hembergerik@csail.mit.edu}
\and Una-May O'Reilly  \at MIT CSAIL, Cambridge, MA, USA \email{unamay@csail.mit.edu}}
%
%
\maketitle

\abstract{
  We investigate training Generative Adversarial Networks,
  GANs, with less data. Subsets of the training dataset can express
  empirical sample diversity while reducing training resource
  requirements, e.g. time and memory. We ask how much data reduction
  impacts generator performance and gauge the additive value of generator ensembles.  In addition to
  considering stand-alone GAN training and ensembles of generator
  models, we also consider reduced data training on an evolutionary GAN training framework named \method. \method  makes GAN training more robust and accurate by exploiting
  overlapping neighborhood based training on a spatial 2D
  grid.  We conduct empirical experiments on \method using  the MNIST and CelebA data sets.
}

\section{Introduction}
\label{sec:introduction}

In Generative Adversarial Network(GAN) training pathologies such as
mode and discriminator collapse can be overcome by using an
evolutionary approach~\cite{wang2018evolutionary, toutouh2019}. In
particular, an evolutionary GAN training method called \horse has been
used for creating robust and accurate generative
models~\cite{toutouh2019}. We work with \method, a descendant of \horse. Per \horse, \method operates on spatially
distributed populations of generators and discriminators. It executes an asynchronous competitive
coevolutionary algorithm on an abstract 2D spatial grid of cells
organized into overlapping Moore neighborhoods. On each cell there is
a subpopulation of generators and the other of discriminators,
aggregated from the cell and its adjacent neighbors. The neural
network models' parameters are updated with stochastic gradient
descent following conventional machine learning. Between training
epochs, the sub-populations are reinitialized by requesting copies of
best neural network models from the cell's neighborhood. This implicit
asynchronous information exchange relies upon overlapping
neighborhoods. In contrast to \horse, only after the final epoch, in \method, the probability weights for each generator ensemble, consisting of a cell and its neighbors, are optimized using an evolutionary strategy.  One model in the ensemble is selected probabilistically, on the basis of the weights, to generate the sample.  

While it has been shown that mixtures of GANs perform well~\cite{arora2017gans}, one drawback of relying upon multiple generators is that it can be resource intensive to train them. A simple
approach to reduce resource use during training is to use less
data. For example, different GANs can be trained on different subsamples of the training data set. The use of less training data reduces the storage requirements,
both disk and RAM while depending on the ensemble of generators to limit possible loss in performance from the reduction of training
data. In the case of \method there is also the potential benefit of the implicit communication that comes from the training on overlapping neighborhoods and updating the cell with the best generator after a training epoch.  This leads to the
following research questions:
\begin{asparaenum}
\item How does the accuracy of generators change in spatially
  distributed grids when the dataset size is decreased?
\item How do ensembles support training with less data in cases where models are trained independently or on a grid with implicit communication?
\end{asparaenum}

The contributions of this chapter are:
\begin{asparaitem}
\item \method, a resource efficient method for evolutionary GAN training,
\item a method for optimizing GAN generator ensemble mixture weights
  via evolutionary strategies
\item analysis of the impact of data size on GAN training on the MNIST and CelebA 
  data sets
\item analysis of the value of ensembling after GAN
  training on subsets of the data.
\end{asparaitem}

We proceed as follows. Notation for this chapter is in
Section~\ref{sec:gan-training}. In Section~\ref{sec:related-work} we
describe related work. The \method is described in
Section~\ref{sec:methods}. Empirical experiments are reported in
Section~\ref{sec:experiments}. Finally conclusions and future work are
in Section~\ref{sec:conclusions}.

\section{General GAN training}
\label{sec:gan-training}

In this study, we adopt the notation similar to~\cite{arora2017generalization,li2017towards}.  Let $\mathcal{G}=\{G_g, g \in \mathcal{U}\}$ and $\mathcal{D}=\{D_d, d \in \mathcal{V}\}$ denote the class of
generators and discriminators, where $G_g$ and $D_d$ are functions
parameterized by $g$ and $d$. $\mathcal{U}, \mathcal{V} \subseteq \mathbb{R}^{p}$
represent the respective parameters space of the generators and
discriminators.  Finally, let $G_*$ be the target unknown distribution
to which we would like to fit our generative model.

Formally, the goal of GAN training is to find parameters $g$ and $d$
in order to optimize the objective function
\begin{equation}
\label{eq:gan-def}
\min_{g\in \mathcal{U}}\max_{d \in \mathcal{V}} \mathcal{L}(g,d)\;, \;\text{where} \nonumber
\end{equation}
\begin{equation}
\mathcal{L}(g,d) = \E_{x\sim G_*}[\phi (D_d(x))] + \E_{x\sim G_g}[\phi(1-D_d(x))]\;,
\end{equation}
and $\phi:[0,1] \to \mathbb{R}$, is a concave \emph{measuring function}.  In
practice, we have access to a finite number of training samples $x_1,
\ldots, x_m \sim G_*$.  Therefore, an empirical version
$\frac{1}{m}\sum_{i=1}^{m} \phi(D_d(x_i))$ is used to estimate
$\E_{x\sim G_*}[\phi (D_d(x))]$. The same also holds for $G_g$.
\section{Related Work}
\label{sec:related-work}

\textbf{Evolutionary Computing and GANs.} 
Competitive coevolutionary algorithms
have adversarial populations (usually two) that simultaneously
evolve~\cite{Hillis1990Coev} population solutions against each other. Unlike classic evolutionary algorithms,
they employ fitness functions that rate solutions relative to their
\emph{opponent} population.  Formally, these algorithms can be
described with a minimax formulation
\cite{Herrmann1999Genetic,Ash2018Reckless} which makes them similar to GANs. 

\textbf{Spatial Coevolutionary Algorithms.}
Spatial (toroidal) coevolution is an effective means of controlling the mixing of adversarial populations in coevolutionary algorithms.  
Five cells per neighborhood (one center and four adjacent cells) are
common~\cite{Husbands1994Distributed}. With this notion of distributed
evolution, each neighborhood can evolve in a different direction and
more diverse points in the search space are explored. Additional investigation into the value of spatial coevolution has been conducted by~\cite{Mitchell2006Coevolutionary,Williams2005Investigating}.

\textbf{Scaling Evolutionary Computing for Machine Learning.}  A team from \textsf{OpenAI}~\cite{salimans2017evolution} applied a simplified version of Natural Evolution Strategies~(NES)~\cite{wierstra2008natural} with a novel communication strategy to a collection of reinforcement learning (RL) benchmark problems. Due to better parallelization over thousand cores, they achieved much faster training times (wall-clock time) than popular RL techniques. Likewise, a team from~\textsf{Uber~AI}~\cite{clune2017uber}
showed that deep convolutional networks with over $4$ million parameters trained with genetic algorithms can also reach results competitive to those trained with \textsf{OpenAI}'s NES and other RL algorithms. \textsf{OpenAI} ran their experiments on a computing cluster of $80$ machines and $1440$ CPU cores~\cite{salimans2017evolution}, whereas \textsf{Uber~AI} employed a range of hundreds to thousands of CPU cores (depending on availability). 
EC-Star~\cite{hodjat2014maintenance} is another example of a large scale evolutionary computation system. By evaluating population individuals only on a small number of training examples per generation, Morse et al.~\cite{morse2016simple} showed that a simple evolutionary algorithm can optimize neural networks of over $1000$ dimensions as effectively as gradient descent algorithms.  FCUBE, see \url{https://flexgp.github.io/FCUBE/} is a cloud-based modeling system that uses genetic programming~\cite{arnaldo2015bring}. 

\textbf{Ensembles - Evolutionary Computation and GANs}
Evolutionary model ensembling has been explored with the aforementioned FCUBE system. FCUBE factors different data splits to cloud instances that model with symbolic regression. These instances draw subsets of variables  and  fitness functions and learn weakly.  After learning the best models are filtered to eliminate the weakest ones and ensemble fusion is used to unify the prediction.  

Bagging applies a weighted average to the outputs of a model set for prediction and assumes that all models use the same input variables. Random forests combine bagging with decision trees that use randomized subsets of the input variables. The ensemble technique of \method has weights that bias probabilistic selection of one model in the ensemble to generate a sample in contrast to these techniques which consider all model outputs and average them.  
There are alternative methods of combining GANs into ensembles. For example, ``self-ensembles'' of GANs were  introduced  by \cite{Wang2016EnsemblesOG} and  are constructed with models based on the same network initialization while training for different numbers of iterations. The same authors introduced also cascade GANs where the part of the training data which is badly modeled by one GAN is redirected to a follow-up GAN.
Other examples include boosting such as \cite{Grover2017BoostedGM} and \cite{Tolstikhin2017AdaGANBG} who present AdaGAN, which adds a new component into a mixture model at each step by running a GAN algorithm on a reweighted sample. MD-GAN \cite{Hardy2018MDGANMG} distributes GANs so that they can be trained over datasets that are spread on multiple workers.  It proposes a novel learning procedure to fit this distributed setup whereas \horse uses conventional gradient-based training and a probabilistic mixture model.  In K-GANS \cite{Ambrogioni2019kGANsEO}  an ensemble of GANs is trained using semi-discrete optimal transport theory.  Quoting the authors, ``each generative network models the transportation map between a point mass (Dirac measure) and the restriction of the data distribution on a tile of a Voronoi tessellation that is defined by the location of the point masses. We iteratively train the generative networks and the point masses until convergence.''  
MGAN~\cite{Hoang2018MGANTG} trains with multiple generators given the specific goal of overcoming mode collapse. They add a classifier to the architecture and use it to specify which generator a sample comes from.  Essentially,  internal samples are created from multiple generators and then one of them is randomly drawn to provide the sample. With the specific aim to provide complete guaranteed mode coverage, \cite{Zhong2019RethinkingGC} constructing the generator mixture with a connection to the multiplicative weights update rule.  

The next section presents \method: a scalable, distributed framework for coevolutionary GAN training with reduced training data use.

\section{Data Reduction in Evolutionary GAN Training}
\label{sec:methods}


This section describes \method which is a spatially distributed
coevolutionary GANs training method in which GANs at each cell are trained by using
subsets of the whole training data set.  The key output of
\method is the best performing ensemble (mixture) of generators.  
First we describe the spatial topology used to evolutionary train GANs in Section~\ref{sec:spat-evol-gan}. Next we present how we subsample the training data in Section~\ref{sec:coev-gans}. 
Then in Section~\ref{sec:weights-evolution} we describe how the final
generator mixture weights are determined. 
Finally, we formalize the \method algorithm in Section~\ref{sec:algorithms}.

\subsection{Overview of \method}
\label{sec:spat-evol-gan}

\method is an extension of \horse and addresses the robust training of
GANs by employing an adversarial arms races between two populations,
one of generators and one of discriminators.  Going forward, we use the
term \textit{adversarial populations} to denote these two populations.
Thus, we define a population of generators $\mathbf{g}=\{g_1, \dots,
g_Z\}$ and a population of discriminators $\mathbf{d}=\{d_1, \dots,
d_Z\}$, where $Z$ is the size of the population.  These two
populations are trained one against the other.  The use of populations
are one source of diversity, that has shown to be adequate to deal
with some of the GAN's training
pathologies~\cite{schmiedlechner2018towards}.

\method defines a toroidal grid.  In each cell, it places a GAN (a
pair generator-discriminator), which is named \textit{center}.  Each
cell has a neighborhood that forms a subpopulations of models:
$\mathbf{g}$ (generators) and $\mathbf{d}$ (discriminators).  The size
of these subpopulations is denoted by $s$.  In this study, \method
uses five-cell Moore neighborhood ($s=5$), i.e., the neighborhoods
include the cell itself (center) and the cells in the \textit{west},
\textit{north}, \textit{east}, and \textit{south}. 

\begin{figure}[!h]
  \includegraphics[width=0.99\textwidth]{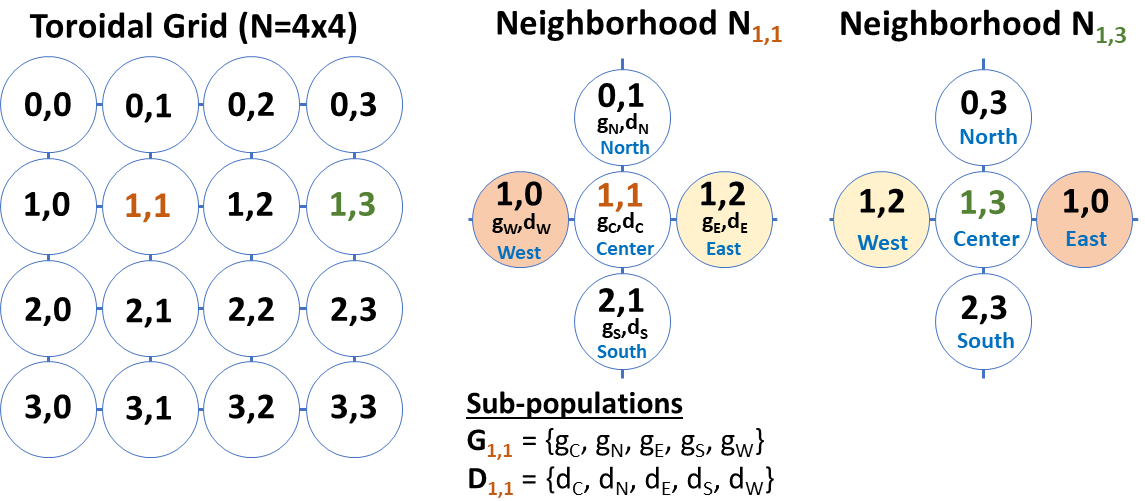}
  \caption{Illustration of overlapping neighborhoods on a toroidal
    grid.  Note how a cell update at N$_{1,2}$ can be communicated to
    N$_{1,1}$ and N$_{1,3}$ when they gather their neighbors. If
    N$_{1,1}$ is then updated with the updated value from N$_{1,2}$
    the value has propagated. If the propagation continues one more
    cell to the left to N$_{1,0}$, the value will come into the range
    of both N$_{0,0}$ and N$_{2,0}$. Propagation runs laterally and
    vertically. We also show an example of a cell's generator and
    discriminator sub-populations (based on its neighborhood) for
    N$_{1,1}$.}
  \label{fig:4x4-nhood}
\end{figure}

For the $k$-th neighborhood in the grid, we refer to the generator in
its center cell by $\mathbf{g}^{k,1}\subset \mathbf{g}$ and the set of
generators in the rest of the neighborhood cells by \method
$\mathbf{g}^{k,2}, \ldots, \mathbf{g}^{k,s}$,
respectively. Furthermore, we denote the union of these sets by
$\mathbf{g}^k = \cup^{s}_{i=1} \mathbf{g}^{k,i} \subseteq
\mathbf{g}$, which represents the $k$th generator neighborhood.  Note
that given a grid size $m\times m$, there are $m^2$
neighborhoods. Fig.~\ref{fig:4x4-nhood} illustrates some examples of
the overlapping neighborhoods on a $4 \times 4$ toroidal grid and how
the sub-populations of each cell are built ($G_{1,1}$ and
$D_{1,1}$). The use of this grid for training the models addresses the
quadratic computational complexity of the basic adversarial
competitions based algorithms.  Without loss of generality, we
consider square grids of $m\times m$ size in this study.

The overlapping neighborhoods define the possible exchange of
information among the different cells during the training process due
to the \textit{selection} and \textit{replacement} operators applied
in coevolutionary algorithms.

\method is built on the following basis: \textit{selection and
  replacement}, \textit{fitness evaluation}, and \textit{reproduction
  based on GAN training}.

\textbf{Selection and Replacement} 
Selection promotes high performing solutions when updating a subpopulation. \method applies \emph{tournament selection} of size $\tau$ to update the center of the cell. 
First, the subpopulations with the updated copies of the neighbors evaluate all GAN generator-discriminator pairs,  then $\tau$ generators and $\tau$  discriminators are randomly picked, 
and the center of the cell is set as the fittest generator and discriminator from the $\tau$ selected ones (lines from \ref{alg:lipi-get-minibatch} to
\ref{alg:lipi-select-center} of~\alg~\ref{alg:lipi}). 
After all GAN training is completed
all models are evaluated again, and the tournament selection is applied to replace the least fit generator and discriminator in the subpopulations with the fittest ones and sets
them as the center of the cell (lines from \ref{alg:fit-eval-1.5} to
\ref{alg:replace-center} of~\alg~\ref{alg:lipi}).

\textbf{Fitness Evaluation}
The search and optimization in evolutionary algorithms are guided by
the evaluation of the \textit{fitness}, 
a measure that evaluates how good a solution is at solving the problem. 
In \method, an adversarial
method, the performance of the model depends on the adversary.  The
performance of a given generator (discriminator) is evaluated in terms
of some loss function $M$. 
\method uses \emph{Binary cross entropy (BCE) loss} (see Equation~\ref{eq:minmax-mut}),
where the model's objective is to minimize the Jensen-Shannon
divergence (JSD) between the real ($p$) and fake ($q$) data
distributions, i.e., $JSD(p\parallel q)$.
In \method, fitness
$\mathcal{L}$ of a model ($g_i\in \mathbf{g}$ or $d_j \in \mathbf{d}$)
is its average performance against all its adversaries. 
\begin{align}
    \label{eq:minmax-mut}
    M^{BCE} &= \frac{1}{2} \E_{x\sim G_g}[log(1-D_d(x))]
  \end{align}

\textbf{Variation - GAN training}
Model variation is done via GAN training, which is applied in order to
update the parameters of the models.  Stochastic Gradient Descent
training performs gradient-based updates on the parameters (network weights)
the models. Moreover, Gaussian-based updates create new learning rate values $n_{\delta}$.

The center generator (discriminator) is trained against a randomly
chosen adversary from the subpopulation of discriminators (generators)
(lines \ref{alg:lipi-rand-d} and \ref{alg:lipi-rand-g} of~\alg~\ref{alg:lipi},
respectively).

\subsection{Dataset Sampling in \method}
\label{sec:coev-gans}

\begin{figure}[!h]
\centering
  \includegraphics[width=0.75\textwidth]{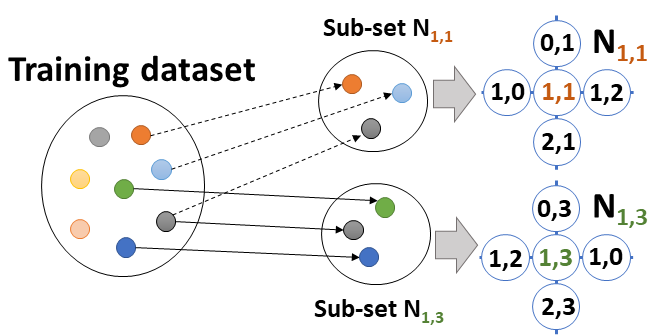}
  \caption{Illustration of how the training dataset is sampled to
    generate training data subsets to train the
    different neighborhoods on the grid (N$_{1,1}$ and N$_{1,3}$).}
  \label{fig:data-sampling}
\end{figure}

Instead of training each sub-population with the whole training
dataset, per \horse, \method applies random sampling with replacement
over the training data to define $m^2$ different subsets (partitions)
of data that will be used as training dataset for each cell (see
Figure~\ref{fig:data-sampling}).  Thus, each cell has its own
\textit{training subset} of data.

\subsection{Evolving Generator Mixture Weights}
\label{sec:weights-evolution} 

\method searches for and returns a mixture of generators composed from
a neighborhood.  The mixture of generators is the fusion of the
different generators in the neighborhood trained by subsets of the
training data set. The selection of the \textit{best} weights that
define mixture ensemble is difficult.  \method applies an
\texttt{ES}-(1+1) algorithm \cite[Algorithm
  2.1]{loshchilov2013surrogate} to evolve a mixture weight vector
$\mathbf{w}$ for each neighborhood in order to optimize the
performance of the fused generative model, see
Algorithm~\ref{alg:mixture-ea}.
  
When using \method for images Fr\'echet Inception Distance (FID)
score~\cite{Martin2017GANs} is used to asses the accuracy of the
generative models.  Note, nothing prevents the use of different
metrics, e.g., inception score.

The $s$-dimensional mixture weight vector $\v{w}$ is defined as follows
\begin{equation}
\footnotesize
\mathbf{g}^*, \v{w}^* =\argmin_{\mathbf{g}^{k}, \v{w}^k: 1\leq k \leq m^2  } \sum_{{g_i \in \mathbf{g}^{k} \\ w_i \in \v{w}^k}} w_i FID_{g_i},
\label{eq:mixture}
\end{equation}
where $w_i$ represents the probability that
a data point comes from the $i$th generator in the neighborhood, with
$\sum_{w_i \in \v{w}^k} w_i = 1$. 

 \subsection{Algorithms of \method}
 \label{sec:algorithms}

\alg~\ref{alg:lipizzaner} formalizes the main steps of \method.  First, it starts the
parallel execution of the training on each cell by initializing their
own learning hyper-parameters (i.e., the \textit{learning rate} and
the mixture weights) and by assigning them their own training
subsets (Lines~\ref{alg:creating-data-sets}
and~\ref{alg:lipi-initialize}).  Then, the training process, see \alg~\ref{alg:lipi},  consists of a loop with two main steps: first, gather the GANs
(neighbors) to build the subpopulations (neighborhood) and, second,
update the center by applying the coevolutionary GANs training
method for all mini-batches in the training subset. 
These steps are repeated $T$ (generations or training epochs) times.
After that, each cell optimizes its mixture weights by applying an
Evolutionary Strategy in order to optimize the performance
of the ensemble defined by the neighborhood, see \alg~\ref{alg:mixture-ea}.
Finally, the best performing ensemble is selected across the
entire grid and returned, including its probabilistic weights, as the
final solution.

\begin{algorithm*}[h!]
	\small
	\caption{\method: 
		In parallel, for each cell, initialize settings then iterate over each generation. Each generation, retrieve neighbor cells to build generator and discriminator sub-populations, evolve generators and discriminators trained with SGD, replace worst with best and update self with best, and finally evolve weights for a neighborhood mixture model.\\
		\textbf{Input:} ~T~: Total generations, ~$E$~: Grid cells, ~$k$~: Neighborhood size,
		~$\theta_{D}$~: Training dataset,
		~$\theta_{p}$~: Sampling size in terms of dataset portion,
		~$\theta_{COEV}$~: Parameters for \nameAlgCell,
		~$\theta_{EA}$: Parameters for \texttt{MixtureEA} \newline
		\textbf{Return:}
		~$n$~: neighborhood, ~$\omega$~: mixture weights
	}\label{alg:lipizzaner}
	\begin{algorithmic}[1]
		
		\ParFor{$c \in E$} \Comment{Asynchronous parallel execution of all cells in grid}
		\State $ds \gets$ getDataSubset($\theta_{D}, \theta_{p}$) \Comment{Creates a sub-set of the dataset}\label{alg:creating-data-sets} 
		\State $n, \omega \gets$ initializeCells($c, k, ds$) \Comment{Initialization of cells}\label{alg:lipi-initialize}
		\For{generation} $\in [0,\dots, \text{T}]$ \Comment{Iterate over generations}
		 \label{alg:copy-neighboor}
         \State $n \gets$ copyNeighbours($c, k$) \Comment{Collect neighbor cells for the subpopulations} \label{alg:copy-neighboor}
		\State $n \gets$ \nameAlgLipiTraining($n, \theta_{COEV}$) \Comment{ Coevolve GANs using \alg~\ref{alg:lipi}}
		\EndFor
		\State $\omega \gets$ \texttt{MixtureEA}($\omega, n, \theta_{EA}$)\Comment{Evolve mixture weights, \alg~\ref{alg:mixture-ea}}
		\EndParFor
		\State \Return $(n, \omega)^*$ \Comment{ Cell with best generator mixture } 
	\end{algorithmic}
\end{algorithm*}

%


\begin{algorithm*}[h!]
	\small
	\caption{\nameAlgLipiTraining:  Select a new neighborhood from the current one. Each mini-batch train discriminators against a randomly drawn generator and generators against a randomly drawn discriminator, using SGD.  Evaluate all against each other, using minimum loss as value to choose best to replace worst and update center. Return this new neighborhood.\newline
		\textbf{Input:}
		~$\tau$~: Tournament size, ~$X$~: Input training dataset,
		~$\beta$~: Mutation probability, ~$n$~: Cell neighborhood subpopulation, ~$ds$~: Sub-set of the training dataset \newline
		\textbf{Return:}
		~$n$~: Cell neighborhood subpopulations
	}\label{alg:lipi}
	\begin{algorithmic}[1]
	\State $\mathbf{B} \gets $ getMiniBatches($ds$) \Comment{ Load mini-batches } \label{alg:lipi-get-minibatch}
	\State $B \gets $ getRandomMiniBatch($\mathbf{B}$) \Comment{ Get a random mini-batch to evaluate GAN pairs }	
	\For{$g, d \in \mathbf{g} \times \mathbf{d}$} \Comment{Evaluate all GAN pairs}
		\State $\mathcal{L}_{g,d} \gets$ evaluate($g, d, B$) \Comment{ Evaluate GAN} \label{alg:fit-eval-1}
		\EndFor
	  \State $\mathbf{g}, \mathbf{d} \gets$ select($n, \tau$) \Comment{Tournament selection with minimum loss($\mathcal{L}$) as fitness } \label{alg:lipi-select-center}
		
		\For{$B \in \mathbf{B}$} \Comment{Loop over batches}
		\State $n_{\delta} \gets$ mutateLearningRate($n_{\delta}, \beta$) \Comment{ Update neighborhood learning rate} 
		\State $d \gets$ getRandomOpponent($\mathbf{d}$) \Comment{ Get uniform random discriminator} \label{alg:lipi-rand-d}
		\For{$g \in \mathbf{g}$} \Comment{Evaluate generators and train with SGD}
		\State $\nabla_{g} \gets$ computeGradient($g, d$) \Comment{ Compute gradient for neighborhood center }
		\State $g \gets$ updateNN($g, \nabla_g, B$) \Comment{ Update with gradient }
		\EndFor
		\State $g \gets$ getRandomOpponent($\mathbf{g}$) \Comment{ Get uniform random generator} \label{alg:lipi-rand-g}
		\For{$d \in \mathbf{d}$} \Comment{Evaluate discriminator and train with SGD}
		\State $\nabla_{d} \gets$ computeGradient($d, g$) \Comment{ Compute gradient for neighborhood center }
		\State $d \gets$ updateNN($d, \nabla_d, B$) \Comment{ Update with gradient }
		\EndFor
		\EndFor
		
		\For{$g, d \in \mathbf{g} \times \mathbf{d}$} \Comment{Evaluate all updated GAN pairs} \label{alg:fit-eval-1.5}
		\State $\mathcal{L}_{g,d} \gets$ evaluate($g, d, B$) \Comment{ Evaluate GAN} \label{alg:fit-eval-2}
		\EndFor
		\State $\mathcal{L}_{g} \gets \min(\mathcal{L}_{\cdot, d})$ \Comment{ Fitness for generator is the average loss value ($\mathcal{L}$)}
		\State $\mathcal{L}_{d} \gets \min(\mathcal{L}_{g,\cdot})$ \Comment{ Fitness for discriminator is the average loss value ($\mathcal{L}$)}    
		\State $n \gets$ replace($n, \mathbf{g}$) \Comment{ Replace the generator with worst loss}
		\State $n \gets$ replace($n, \mathbf{d}$) \Comment{ Replace the discriminator worst loss}
                \State $n \gets$ setCenterIndividuals($n$) \Comment{Best generator and discriminator are placed in the center} \label{alg:replace-center}
		\State \Return $n$
	\end{algorithmic}
\end{algorithm*}
\clearpage

\begin{algorithm*}[h!]
	\small
	\caption{\nameAlgMixture: Evolve mixture weights $\omega$ with a ES-(1+1). \newline
		\textbf{Input:}
		~$GT$~: Total generations to evolve the weights, ~$\mu$~: Mutation rate , ~$n$~: Cell neighborhood subpopulation, $\omega$~: Mixture weights \newline
		\textbf{Return:}
		~$\omega$~: mixture weights
	}\label{alg:mixture-ea}
	\begin{algorithmic}[1]    
	\For{generation} $\in [0,\dots, \text{GT}]$  \Comment{Loop over generations}
		
                \State $\omega' \gets $ mutate$(\omega, \mu)$ \Comment{Gaussian mutation of mixture weights }
		\State $\omega_f' \gets $ evaluateMixture($\omega', n$) \Comment{Evaluate generator mixture score, e.g. FID for images}
                \If {$\omega_f' < \omega_f$} \Comment{ Replace if new mixture weights are better}
        		\State $\omega \gets \omega'$ \Comment{ Update mixture weights}
                \EndIf
         \EndFor
		\State \Return $\omega$
	\end{algorithmic}
\end{algorithm*}

Section~\ref{sec:experiments} next presents results regarding the
question of how \method performs given data reduction with the support of ensembles.

\newcommand{\ensemble}{mixture\xspace}

\vspace{-0.2cm}
\section{Experimental Analysis}
\label{sec:experiments}
\vspace{-0.2cm}

In this section we proceed experimentally. We use Section~\ref{sec:experimental-setup} to present our experimental setup. We then investigate the following research questions: 
\begin{description}
\item[\textbf{RQ1:}] How robust are spatially distributed grids when training with less of the dataset?
\item[\textbf{RQ2:}]  Given the use of ensembles, if we reduce the data quantity at each cell, at what point will the ensemble fail to fuse the resulting models towards achieving sufficient accuracy?
\end{description}

\vspace{-0.2cm}
\subsection{Experimental Setup}
\label{sec:experimental-setup}
\vspace{-0.2cm}

We use two common image datasets from the GAN literature: MNIST~\cite{lecun1998mnist} and CelebA~\cite{liu2015faceattributes}. 
MNIST has been widely used and it consist of low dimensional handwritten digits images.  
The larger CelebA dataset contains more than 200,000 images of faces.  To obtain an absolute measure of model accuracy, we draw fake image samples from the generative models computed and score them with Frechet inception distance (FID)~\cite{heusel2017gans}.  FID score is a black box, discriminator-independent, metric and expresses image similarity to the samples used in training. 

The process of sampling the data is independent for each cell of the grid and it consist on randomly selecting different mini-batches of the training dataset.  
In the context of a grid, given grid size, there is an expectation that every sample will be drawn at least once.  This can be considered 100\% coverage, \textit{over the grid}, though not at any cell.   When the subset size is lower and/or the grid is smaller, this expected coverage of the complete dataset is nonetheless higher than that of a subset drawn for a single GAN trained independently of others.

For a fixed budget of training samples, when a GAN is trained with a larger dataset and the batch size of a smaller dataset is maintained, the gradient is estimated more often because there are more mini-batches per generation. (Given the standard terminology that an epoch is one forward pass and one backward pass of all the training examples, one epoch is one generation.)  In contrast, in the same circumstances, if the number of mini-batches is held constant, and the mini-batch size increased, we incur a cost increase in RAM to store the mini-batch and the gradient is estimated on better information but less frequently. To date, there is no clear well-founded procedure or even a heuristic for setting mini-batch size. 


We place all experiments on equal footing by training them with the same budget of mini-batches while keeping mini-batch size, i.e. the number of examples per mini-batch, constant. We experimentally vary the training set size per cell or GAN and adjust the number of generations to arrive at the mini-batch budget. See Equation~\ref{eq:batches}.

\begin{equation}
batches\_to\_train = \frac{training\_dataset\_size}{mini-batch\_size} \times data\_portion \times generations
\label{eq:batches}
\end{equation}

For example, given a budget of $1.2\times 10^5$ mini-batches and a mini-batch size of 100, when the training set size per cell is 60000, there will be 600 mini-batches per generation. We therefore train for 200 generations to reach the $1.2\times 10^5$ mini-batches budget. When we reduce the training set size to 30000 ($50\%$), there will be only 300 mini-batches per generation so we train for 400 generations to reach the training budget of $1.2\times 10^5$ mini-batches.

Considering the dataset sizes in terms of images (60,000 in MNIST and 202,599 in CelebA), a constant batch size of 100 (Table~\ref{tab:exp-data-partition}), and the relative training data subset size, we provide the number of generations executed in Table~\ref{tab:algs-configuration}.  The total number of batches used to train MNIST is 1.20$\times 10^5$ and CelebA is 31.66$\times 10^3$ when training with the 100\% of the data.  

\begin{table}[h!]
\setlength{\tabcolsep}{6pt}
	\centering
	\small
  \caption{\small Batches and generations used in experimental comparisons under equalization to the same computational budget (expressed as batches).}
  \label{tab:algs-configuration}
  \begin{tabular}{lrrrr}
    
    \textbf{Portion of data} & \textbf{100\%} & \textbf{75\%} & \textbf{50\%} & \textbf{25\%} \\
    	\rowcolor{GAINSBORO}
		\multicolumn{5}{c}{\textit{MNIST (Computation budget = 1.20$\times 10^5$)}} \\ 
    \textbf{Number of mini-batches}  & 600 & 450 & 300 & 150 \\
    \rowcolor{ALICEBLUE}
    \textbf{Number of generations}  & \textit{200} & \textit{267} & \textit{400} & \textit{800}  \\ 
    	\rowcolor{GAINSBORO}
		\multicolumn{5}{c}{\textit{CelebA (Computation budget = 31.66$\times 10^3$)}} \\ 
	    \textbf{Number of mini-batches}  & 1583 & 1187 & 792 & 396 \\
    \rowcolor{ALICEBLUE}
    \textbf{Number of generations}  & \textit{20} & \textit{27} & \textit{40} & \textit{80}  \\ 
  \end{tabular}
\end{table}

In this analysis, we compare \method by using different grid sizes (\xGrid{4} and \xGrid{5} for MNIST and \xGrid{3} for CelebA) with a \textit{Single GAN} training method. 
These different grid sizes allow us to explore the performance of \method according to different degrees of cell overlap. 
The datasets selected, MNIST and CelebA, represent different challenges for GANs training due to: 
first, the size of each sample of MNIST (vector of 784 real numbers) is smaller than the same of CelebA (vector of 12,288 real numbers); 
second, MNIST dataset has fewer number of samples than CelebA, and third, the size of the models (generator-discriminator) are much larger for CelebA generation than for MNIST. 
This makes the computational resources required to address CelebA higher than for MNIST. 
Thus, we have defined our experimental analysis taking into account different overlapping patterns and datasets, but also the computational resources available.

All these methods are configured according to the parameterization shown in Table~\ref{tab:exp-data-partition}.  
In order to extend our analysis, we apply a \textit{bootstrapping} procedure to compare the \textit{Single GANs} with \method. Therefore, we randomly generated 30~populations (grids) of 16 and 25 generators from the 30 generators computed by using \textit{Single GAN} method. Then, we compute their FIDs and create \ensemble{s} to compare these results against \method \xGrid{4} and \method \xGrid{5}, respectively. We name these variants \textit{Bootstrap} \xGrid{4} and \textit{Bootstrap} \xGrid{5}.


\begin{table}[h!]
\setlength{\belowcaptionskip}{-5pt}
\setlength{\tabcolsep}{6pt}
	\centering
	\small
	\caption{\small Setup for experiments conducted with \method on MNIST and CelebA}
	\label{tab:exp-data-partition}
	\begin{tabular}{l|l|l}
		\textbf{Parameter} & \textbf{MNIST} & \textbf{CelebA} \\
		\hline
                \rowcolor{GAINSBORO}
		\multicolumn{3}{c}{\textit{Coevolutionary settings}} \\ 
		Generations & \multicolumn{2}{c}{\textit{See Table \ref{tab:algs-configuration}}} \\
                \rowcolor{ALICEBLUE}
		Population size per cell & 1 & 1 \\ 
		Tournament size & 2 & 2\\ 
                \rowcolor{ALICEBLUE}
		Grid size & \xGrid{1}, \xGrid{4}, and \xGrid{5} & \xGrid{3} \\
		\rowcolor{GAINSBORO}
		\multicolumn{3}{c}{\textit{Mixture evolution}} \\ 
		Mixture mutation scale & 0.01 & 0.01 \\ 
		Generations & 5000 & 5000 \\ 
                \rowcolor{GAINSBORO}
		\multicolumn{3}{c}{\textit{Hyper-parameter mutation}} \\ 
		Optimizer & Adam & Adam \\ 
                \rowcolor{ALICEBLUE}
		Initial learning rate & 0.0002 & 0.00005 \\ 
		Mutation rate & 0.0001 & 0.0001 \\ 
                \rowcolor{ALICEBLUE}
		Mutation probability & 0.5 & 0.5 \\ 
                \rowcolor{GAINSBORO}
                \multicolumn{3}{c}{\textit{Network topology}} \\ 
		Network type  & MLP & DCGAN \\ 
                \rowcolor{ALICEBLUE}
		Input neurons & 64 & 100 \\ 
		Number of hidden layers & 2 & 4 \\ 
                \rowcolor{ALICEBLUE}
		Neurons per hidden layer & 256 & $16,384 - 131,072$ \\ 
		Output neurons  & 784 & $64 \times 64 \times 3$ \\ 
                \rowcolor{ALICEBLUE}
		Activation function & $tanh$ & $tanh$ \\ 
                \rowcolor{GAINSBORO}
		\multicolumn{3}{c}{\textit{Training settings}} \\ 
		Mini-batch size  & 100 & 128 \\ 
	\end{tabular}
\end{table}

All methods have been implemented in \texttt{Python3} and
\texttt{pytorch}\footnote{Pytorch Website -
  \texttt{https://pytorch.es/}}. 
The experiments are performed on a cloud that provides 16~Intel Cascade Lake cores up to 3.8 GHz with 64~GB RAM and a GPU which are either NVIDIA Tesla P4 GPU with 8~GB RAM or NVIDIA Tesla P100 GPU with 16~GB RAM.  
All implementations use the same \texttt{Python} libraries and versions to minimize computational differences that could arise from using the cloud.

\subsection{Research Question 1: How does the accuracy of generators change in spatially distributed grids when the dataset size is decreased? }\label{sec:exp-rq1}

\begin{figure}[h!]
\vspace{-0.5cm}
\subfloat[\scriptsize{MNIST 25\%}]
         {\includegraphics[width=0.4\textwidth]{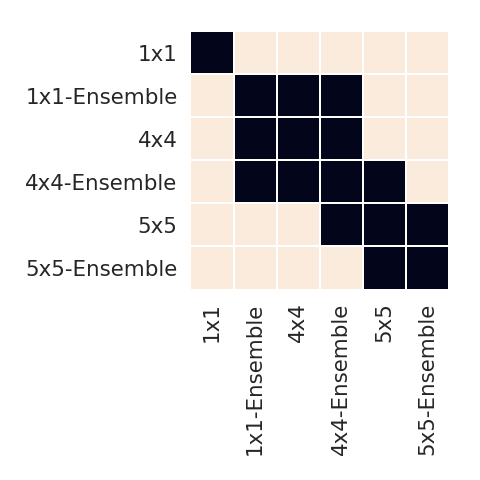}\label{fig:stats-mnist-25}}
\subfloat[\scriptsize{MNIST 50\%}]
         {\includegraphics[width=0.4\textwidth]{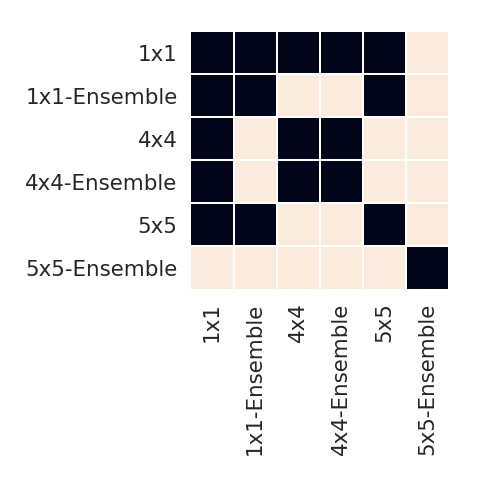}\label{fig:stats-mnist-50}}
         
\subfloat[\scriptsize{MNIST 75\%}]
         {\includegraphics[width=0.4\textwidth]{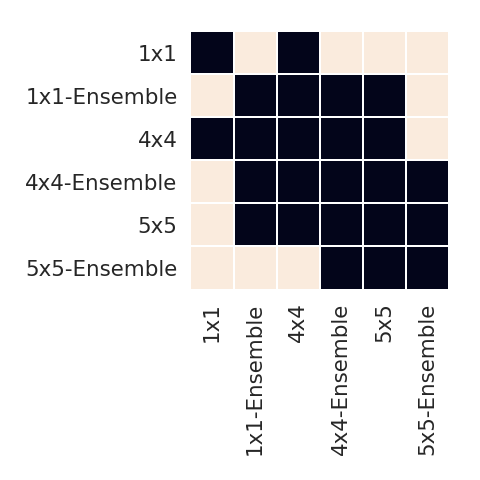}\label{fig:stats-mnist-75}}
\subfloat[\scriptsize{MNIST 100\%}]
         {\includegraphics[width=0.4\textwidth]{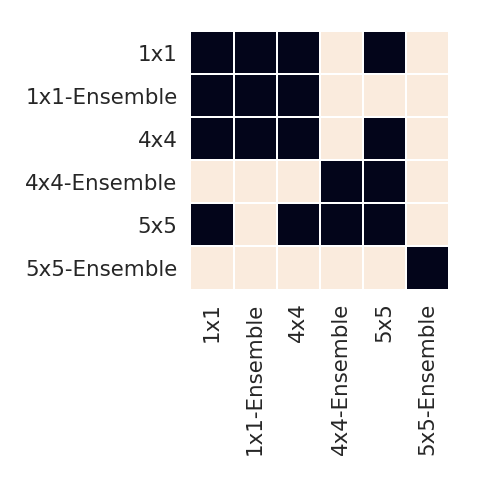}\label{fig:stats-mnist-100}}

\caption{\small{Statistical analysis comparing the same amount of data applying different methods. Black indicates no significance with $\alpha=0.01$}} 
  \label{fig:statistical-tests-data}
  \vspace{-0.3cm}
\end{figure}

We first establish a non-grid baseline by training a single GAN with decreasing amounts of data and examining the resulting FID scores, see Table~\ref{tbl:mean-fids}, and Fig.~\ref{fig:statistical-tests-data} for pairwise statistical significance with a Wilcoxon Rank Sum test with $\alpha=0.01$.   What we see is obvious, FID score increases (performance worsens) as the GAN is trained on less data. In 30 runs of single GAN training on $25\%$ of the data, the mean FID score is very high: 574.6 while the standard deviation of FID score  is $51.3\%$ including the best FID score of $35.1$.  When the data subset  is doubled to $50\%$, the mean FID score drops to $71.2$ but the observed standard deviation is higher ($104.6\%$). The best FID score falls to $30.1$.  Mean FID score improves with $75\%$ of the data significantly (from $71.2$ to $39.8$) but minimally in terms of the best FID score (30.1 vs 30.2), see Table~\ref{tbl:best-fids}.  A marked decrease in standard deviation (104.6\% to  12.4\%) occurs.  In all cases, the smaller training subsets do not match the performance when training with 100\% of the data where the best FID score is $27.4$  and the mean FID score is $38.8$, see Table~\ref{tbl:mean-fids} and Table~\ref{tbl:best-fids}.   These results are straight forwardly explained by smaller quantities of data failing to sufficiently cover the latent distribution.

\begin{table}
\setlength{\tabcolsep}{2pt}
	\centering
	\small
	\caption{\small Mean($\pm$std) of the \textit{best FID in the grid} for 30 independent runs}
	\label{tbl:mean-fids}
\begin{tabular}{ll|rrrr}
\rowcolor{GAINSBORO}
\textbf{Dataset} &  \textbf{Variant} &   \textbf{25\%} &  \textbf{50\%} &  \textbf{75\%} &  \textbf{100\%}\\
\hline
MNIST & Single GAN & 574.6$\pm$51.3\%  &  71.2$\pm$104.6\%  &  39.8$\pm$12.4\%  &  38.8$\pm$17.0\%   \\ 
\rowcolor{ALICEBLUE}
MNIST & Single GAN Ensemble & 44.2$\pm$9.5\%  &  35.4$\pm$8.0\%  &  34.4$\pm$10.0\%  &  38.6$\pm$12.3\%   \\ 
\hline
MNIST & Bootstrap \xGrid{4} &  578.0$\pm$12.3\%  &  73.5$\pm$25.3\%  &  39.7$\pm$3.0\%  &  38.9$\pm$4.0\%   \\  
\rowcolor{ALICEBLUE}
MNIST & Bootstrap \xGrid{4} Ensemble & 44.2$\pm$9.0\%  &  35.4$\pm$5.1\%  &  34.5$\pm$6.6\%  &  35.9$\pm$7.4\%   \\ 
\hline
MNIST & \method \xGrid{4} & 47.0$\pm$19.2\%  &  42.0$\pm$16.5\%  &  36.5$\pm$19.2\%  &  37.3$\pm$15.1\%   \\ 
\rowcolor{ALICEBLUE}
MNIST & \method \xGrid{4} Ensemble & 44.1$\pm$21.9\%  &  40.5$\pm$15.4\%  &  33.6$\pm$16.7\%  &  30.7$\pm$17.3\%   \\ 
\hline
MNIST & Bootstrap \xGrid{5} & 573.2$\pm$9.5\%  &  74.8$\pm$22.6\%  &  39.7$\pm$2.3\%  &  38.7$\pm$3.3\%   \\ 
\rowcolor{ALICEBLUE}
MNIST & Bootstrap \xGrid{5} Ensemble & 43.3$\pm$8.3\%  &  33.3$\pm$6.3\%  &  33.0$\pm$5.4\%  &  34.6$\pm$6.7\%   \\ 
\hline
MNIST & \method \xGrid{5} & 39.9$\pm$15.6\%  &  34.4$\pm$9.1\%  &  32.9$\pm$14.2\%  &  34.3$\pm$19.9\%   \\ 
\rowcolor{ALICEBLUE}
MNIST & \method \xGrid{5} Ensemble & 36.2$\pm$16.9\%  &  31.8$\pm$12.6\%  &  30.1$\pm$15.9\%  &  26.3$\pm$16.7\%   \\ 
\hline
\hline
CelebA & \method \xGrid{3} & 51.9$\pm$29.8\%  &  50.3$\pm$25.7\%  &  51.3$\pm$26.6\%  &  46.5$\pm$7.3\%   \\
\rowcolor{ALICEBLUE}
CelebA & \method \xGrid{3} Ensemble & 49.1$\pm$1.4\%  &  44.7$\pm$6.8\%  &  40.3$\pm$3.9\%  &  43.2$\pm$0.9\%   \\ 
\end{tabular}
\end{table}

We can now consider competitive coevolutionary grid-trained GANs where
there is one GAN per cell, the best of that cell's training, at the end of
execution (see Table~\ref{tbl:mean-fids}). This data allows us to isolate the value of the
evolutionary training's communication in contrast to
\begin{inparaenum}[\itshape a)]
\item the independently trained GANs we previously evaluated and
  (below)
\item the performance impact of ensemble.
\end{inparaenum}
Recall that the overlapping neighborhoods facilitate signal
propagation. GANs which perform well in one neighborhood migrate
to their adjacent and overlapping neighborhoods in a form of
communication.  The grid-trained GANs achieve better FID scores, given
the same training budget, than independently trained GANs~\cite{toutouh2019}.  In the
case of a \xGrid{4} grid, the experimental mean FID score is 37.3 with
a standard deviation of 15.1\% and the best generator has a FID score
of 26.4.  The improvement over independently trained GANs is present
with the \xGrid{5} grid, where the experimental mean FID score is 34.3
with a standard deviation of 19.9\% and the best generator has a FID
score of 20.8. One possible explanation is that the communication
indirectly leads to a mixing of the data subsamples (that are drawn
independently and with replacement) that effectively improves the coverage of the data.

Fig.~\ref{fig:statistical-tests-data} illustrates the statistical analysis of different methods evaluated here when using the same amount of data. 
When using the smallest training datasets (MNIST~25\%), the use of \method with larger grids and allow significant improvements of the results. 
The results provided by \method are also improved when the optimization of the mixture weights is applied.   
With larger datasets (MNIST~75\%), \method \xGrid{4} provides results as competitive as the same with \xGrid{5}. Something similar is observed when the whole data is used to rain the GANs.

\begin{figure}[h!]
\centering
\subfloat[\scriptsize{Single GAN}] 
         {\includegraphics[width=0.3\textwidth]{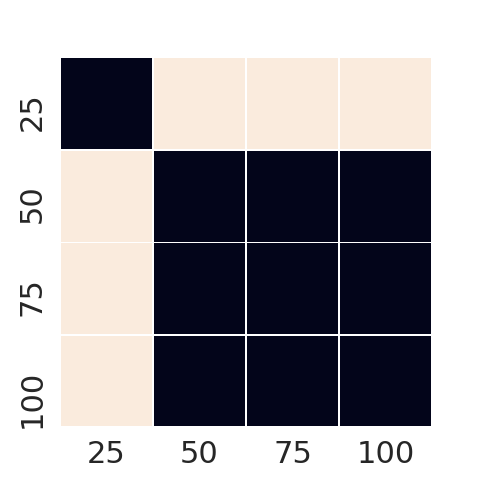}\label{fig:stats-mnist-1x1}} 
         \hspace{0.5cm}
\subfloat[\scriptsize{\xGrid{4}}]
         {\includegraphics[width=0.3\textwidth]{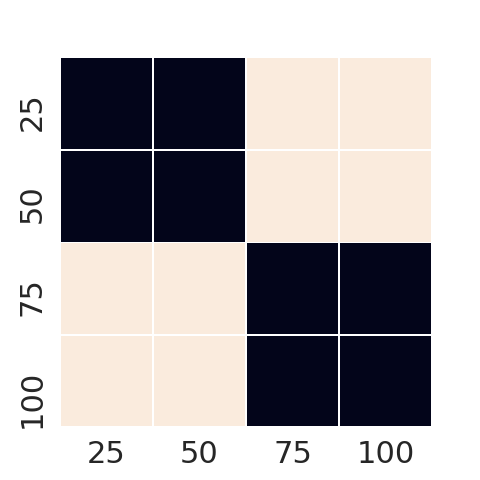}\label{fig:stats-mnist-4x4}}
         \hspace{0.5cm}
\subfloat[\scriptsize{\xGrid{5}}]
         {\includegraphics[width=0.3\textwidth]{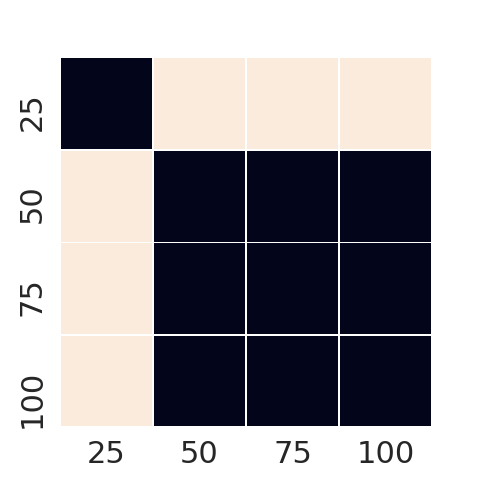}\label{fig:stats-mnist-5x5}}
         
\subfloat[\scriptsize{Single GAN Ensemble}]
         {\includegraphics[width=0.3\textwidth]{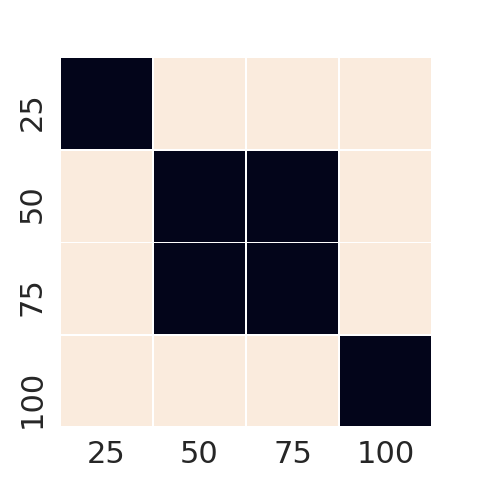}\label{fig:stats-mnist-1x1-Ensemble}}
         \hspace{0.5cm}
\subfloat[\scriptsize{\xGrid{4} Ensemble}]
         {\includegraphics[width=0.3\textwidth]{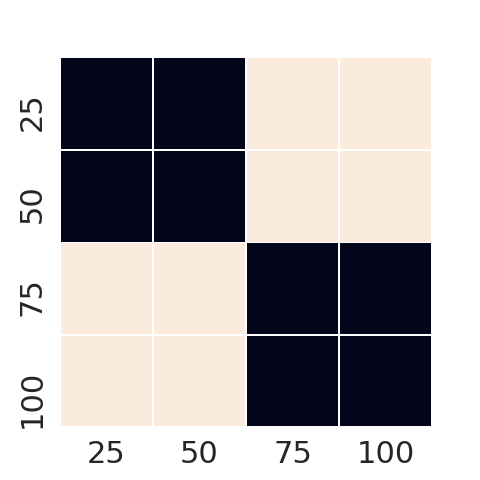}\label{fig:stats-mnist-4x4-Ensemble}}
         \hspace{0.5cm}
\subfloat[\scriptsize{\xGrid{5} Ensemble}]
         {\includegraphics[width=0.3\textwidth]{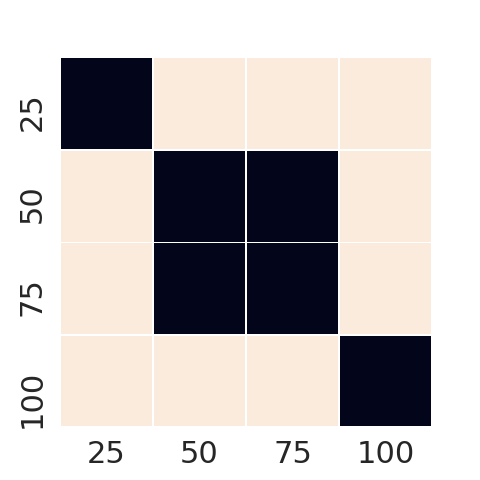}\label{fig:stats-mnist-5x5-Ensemble}}

\caption{\small{Statistical analysis comparing the same method with different size of data for MNIST experiments. Black indicates no significance with $\alpha=0.01$}. Figures (b), (c), (e), and (f) illustrate the results of \method and their \ensemble{s}. Figures (a) and (d) shows the results of bootstrapping and its \ensemble  } 
  \label{fig:statistical-tests-method}
\end{figure}

Fig.~\ref{fig:statistical-tests-method} shows illustrates the statistical analysis of the impact on the performance of the methods analyzed here when using different size of training data. 
\method \xGrid{4} provides similar results when reducing the training data in 25\% (i.e., for MNIST~100\% and MNIST~75\%). 
When the grid size increases, i.e., \method \xGrid{5}, the use o f the training datasets with the half of the data or larger does not show statistical differences in the results. 
However, the application of the \ensemble{s} drives the results with the 100\% of the data to be the most competitive ones.

\begin{table}
\setlength{\tabcolsep}{6pt}
	\centering
	\small
	\caption{\small Min(best) of the \textit{best FID in the grid} for 30 independent runs for different data diets}
	\label{tbl:best-fids}
\begin{tabular}{ll|rrrr}
  \rowcolor{GAINSBORO}
  & & \multicolumn{4}{c}{\textbf{Data diet}} \\
  \rowcolor{GAINSBORO}
\textbf{Dataset} & \textbf{Variant} &   \textbf{25\%} &  \textbf{50\%} &  \textbf{75\%} &  \textbf{100\%}\\
\hline
MNIST & Single GAN & 35.1  &  30.1  &  30.2  &  27.4   \\ 
\rowcolor{ALICEBLUE}
MNIST & Single GAN Ensemble &  33.7  &  30.0  &  26.9  &  27.1   \\ 
\hline
MNIST & Bootstrap \xGrid{4} & 395.5  &  47.0  &  37.0  &  36.1   \\ 
\rowcolor{ALICEBLUE}
MNIST & Bootstrap \xGrid{4} Ensemble & 33.6  &  32.7  &  30.7  &  28.0   \\ 
\hline
MNIST & \method \xGrid{4} & 31.8  &  28.8  &  27.1  &  26.4   \\ 
\rowcolor{ALICEBLUE}
MNIST & \method \xGrid{4} Ensemble & 26.5  &  28.1  &  24.6  &  21.1   \\ 
\hline
MNIST & Bootstrap \xGrid{5} &  440.8  &  48.0  &  37.6  &  34.9   \\ 
\rowcolor{ALICEBLUE}
MNIST & Bootstrap \xGrid{5} Ensemble & 34.8  &  28.1  &  28.0  &  29.8   \\ 
\hline
MNIST & \method \xGrid{5} & 30.5  &  26.8  &  27.3  &  26.3   \\ 
\rowcolor{ALICEBLUE}
MNIST & \method \xGrid{5} Ensemble & 26.3  &  21.9  &  21.2  &  20.8   \\ 
\hline
\hline
CelebA & \method \xGrid{3} & 39.3  &  39.0  &  39.4  &  42.0   \\ 
\rowcolor{ALICEBLUE}
CelebA & \method \xGrid{3} Ensemble & 48.3  &  42.4  &  38.9  &  42.7 \\
\end{tabular}
\end{table}

Scrutinizing Table~\ref{tbl:mean-fids},
Fig.~\ref{fig:statistical-tests-data} and
Fig.~\ref{fig:statistical-tests-method} indicate that \method on a
large grid, \xGrid{5}, performs among the best for all training data
sizes. Furthermore, the \method \xGrid{4} can perform well with
reduced training data size. The CelebA results indicate that \method
has similar response to data dieting on another data set. 
Future work on applying \method to address CelebA with larger grid sizes will allow us to confirm this statement. 
These results lend support to the hypothesis that communication during
training can accelerate the impact of small training datasets. We can
answer affirmatively, for the datasets we examined, for \method.

\subsection{Research Question~2: Given we use ensembles, if we reduce the data quantity at each cell, at what point will the ensemble fail to unify the resulting models towards achieving high accuracy?}\label{sec:exp-rq2}

We can now consider two directions of inquiry, their order not being
important, in the context of the evolutionary GAN training that occurs
on a grid.  As well, each cell's neighborhood on the grid defines a
sub-population of generators and a sub-population of discriminators
which are trained against each other. This naturally suggests the
neighborhood to be the set of GANs in each ensemble of the grid, that
is, the fusing of each center cell's generator with its North, South,
West and East cell neighbors. We therefore can compare the impact of a
neighborhood-based ensemble to each cell's FID score.

We thus isolate communication from grid-based ensembles and we also
isolate co-trained ensembles from ensembles arising from independently
trained GANs. For tabular results see Table~\ref{tbl:mean-fids}.  To
measure the impact of ensembles in this context, i.e. independently
trained GANs, we sample sets of 5 GANS from the 30 different training
runs and train an \ensemble. We formulate experiments with different
portions of the training dataset (i.e., 25\%, 50\%, 75\%, and
100\%. of the samples). We train GANs individually (non-population
based) with the same training algorithm as when we train the GANs
within \horse.

We first measure the improvement of a given method $m$ ($\Delta(m)$) attributable to using
\ensemble{s} ($m_{ensemble}$). 
This metric is evaluated as a percentage in terms of the difference between the average FID of $m$, $\overline{FID(m)}$ and the average FID of the same method when applying the \ensemble{s} $\overline{FID(m_{ensemble})}$, see Equation~\ref{eq:improvement}. 

\begin{equation}
\vspace{0.5cm}
\Delta(m) = \frac{\overline{FID(m)} - \overline{FID(m_{ensemble})}}{\overline{FID(m)}}\ \%
\label{eq:improvement}
\vspace{0.5cm}
\end{equation}

We expect these results to be consistent with
\cite{arora2017gans} who observed an advantage with mixtures of
generators.  We start with the sets of generators obtained from the 30
runs of independent training for each data subset. For each data
subset's set, we optimize the weights of 5 generators randomly drawn
from it via ES-(1+1) for 5,000 generations, and report the mean, min
and std of FID scores for 30 independent draws.  We see
statistically significant improvements for some subsets of data (25\%
and 75\%), see Table~\ref{tbl:ratios-fids}.  The mean ensemble FID
score with 25\% subsets is 44.2 versus the single generator's mean FID
score of 574.6, an improvement of 92.3\%.  The improvement is
diminishes at 50\% to 50.2\% and again at 75\% to 13.7\% and finally
only 0.7\% for 100\%. When all the data is used for training, the
least improvement but still an improvement is observed.

\begin{table}
\vspace{0.3cm}
\setlength{\tabcolsep}{6pt}
	\centering
	\small
	\caption{\small Mean FID improvement $\Delta(m)$ by weighted ensembles for 30 independent runs (Eq.~\ref{eq:improvement}) for different data diets}
	\label{tbl:ratios-fids}
\begin{tabular}{ll|rrrr}
  \rowcolor{GAINSBORO}
  & & \multicolumn{4}{c}{\textbf{Data diet}} \\
\rowcolor{GAINSBORO}
\textbf{Data set} & \textbf{Variant} &   \textbf{25\%} &  \textbf{50\%} &  \textbf{75\%} &  \textbf{100\%}\\
\hline
MNIST & Single GAN &  92.3\%  &  50.2\%  &  13.7\%  &  0.7\% \\
\rowcolor{ALICEBLUE}
MNIST & Bootstrap \xGrid{4} & 92.4\%  &  51.8\%  &  13.1\%  &  7.6\%   \\ 
MNIST & \method \xGrid{4} & 6.2\%  &  3.6\%  &  8.0\%  &  17.5\%  \\
\rowcolor{ALICEBLUE}
MNIST & Bootstrap \xGrid{5} &  92.4\%  &  55.5\%  &  16.9\%  &  10.6\%   \\ 
MNIST & \method \xGrid{5} & 9.3\%  &  7.4\%  &  8.5\%  &  23.5\%   \\ 
\hline
\hline
CelebA & \method \xGrid{3} & 5.3\%  &  11.1\%  &  21.5\%  &  7.1\%   \\ 
\end{tabular}
\vspace{0.3cm}
\end{table}

These results can be anticipated because different subsets were used
in training and the fusion of the generators. An interesting note is
that \method almost has an inverse progression of \ensemble effect,
see Table~\ref{tbl:ratios-fids}, with the \ensemble improving the
performance of \method more the more training data is available.

Moreover, we study the capacity of the generative models created
by using the fusion method presented in
Section~\ref{sec:weights-evolution} (see \alg~\ref{alg:mixture-ea})
from GANs individually trained. In Table~\ref{tbl:best-mean-fids} and
Table~\ref{tbl:mean-mean-fids} the best and the mean FIDs of each
generator is shown. 
The FIDs improves as the training data size increases.  
This again highlights the improvement of the accuracy on the generated samples for the separate generator with more training data.

\begin{table}
\vspace{0.3cm}
\setlength{\tabcolsep}{6pt}
	\centering
	\small
	\caption{\small Min(best) of the \textit{mean FID in the grid} for 30 independent runs for different data diets}
	\label{tbl:best-mean-fids}
\begin{tabular}{ll|rrrr}
  \rowcolor{GAINSBORO}
  & & \multicolumn{4}{c}{\textbf{Data diet}} \\
\rowcolor{GAINSBORO}
\textbf{Dataset} & \textbf{Variant} &   \textbf{25\%} &  \textbf{50\%} &  \textbf{75\%} &  \textbf{100\%}\\
\hline
MNIST & Boostrap \xGrid{4} & 395.5  & 47.0  & 37.0  & 36.1  \\ 
\rowcolor{ALICEBLUE}
MNIST & \method \xGrid{4} & 38.3  &  33.9  &  32.9  &  28.6   \\ 
MNIST & Bootstrap \xGrid{5} & 440.8  & 48.0  & 37.6  & 34.9   \\ 
\rowcolor{ALICEBLUE}
MNIST & \method \xGrid{5} & 34.0  &  29.3  &  30.3  &  27.2   \\ 
\hline
\hline
CelebA & \method \xGrid{3} & 58.2  &  58.6	  &  59.4  &  49.0   \\ 
\end{tabular}
\vspace{0.3cm}
\end{table}

\begin{table}
\setlength{\tabcolsep}{3pt}
	\centering
	\small
	\caption{\small Mean($\pm$std) of the \textit{mean FID in the grid} for 30 independent runs for different data diets}
	\label{tbl:mean-mean-fids}
\begin{tabular}{ll|rrrr}
  \rowcolor{GAINSBORO}
  & & \multicolumn{4}{c}{\textbf{Data diet}} \\
\rowcolor{GAINSBORO}
\textbf{Dataset} & \textbf{Variant} &   \textbf{25\%} &  \textbf{50\%} &  \textbf{75\%} &  \textbf{100\%}\\
\hline
MNIST & Bootstrap\xGrid{4} & 578.0$\pm$12.3\% &  73.5$\pm$25.3\% &  39.7$\pm$3.0\% &  38.9$\pm$4.0\%  \\ 
\rowcolor{ALICEBLUE}
MNIST & \method \xGrid{4} & 55.3$\pm$21.4\%  &  49.7$\pm$16.9\%  &  43.4$\pm$14.8\%    &  40.4$\pm$20.0\%   \\ 
MNIST & Bootstrap \xGrid{5} &  573.2$\pm$9.5\% & 74.8$\pm$22.6\% & 39.7$\pm$2.3\% & 38.7$\pm$3.3\% \\
\rowcolor{ALICEBLUE}
MNIST & \method \xGrid{5} & 46.3$\pm$16.7\%  &  37.0$\pm$15.2\%  &  38.7$\pm$12.0\%  &  34.2$\pm$14.1\%   \\ 
\hline
\hline
CelebA & \method \xGrid{3} & 64.4$\pm$9.8\% &  60.9$\pm$3.8\%& 64.5$\pm$7.3  \% &   51.6$\pm$4.6\%  \\ 
\end{tabular}
\end{table}

\begin{figure}[h!]
\centering
\subfloat[\scriptsize{25\% - Uniform}]
         {\includegraphics[width=0.23\textwidth]{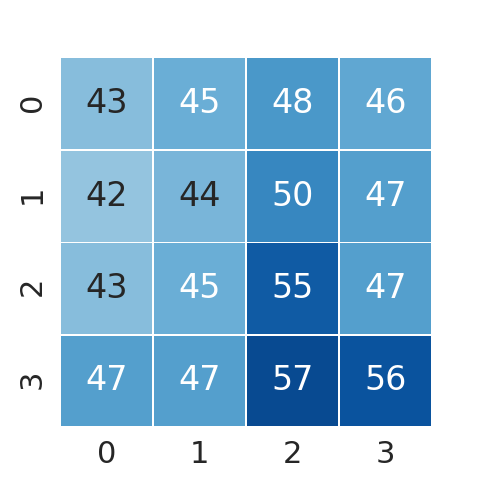}\label{fig:heatmpa-mnist-4x4-025-unif}}\hspace{0.2cm}
\subfloat[\scriptsize{50\% - Uniform}]
         {\includegraphics[width=0.23\textwidth]{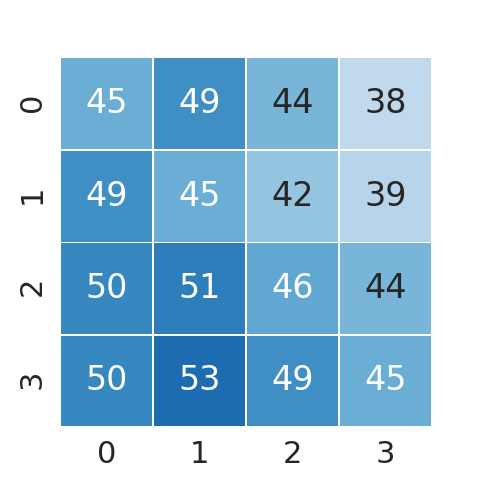}\label{fig:heatmpa-mnist-4x4-050-unif}}\hspace{0.2cm}
\subfloat[\scriptsize{75\% - Uniform }]
         {\includegraphics[width=0.23\textwidth]{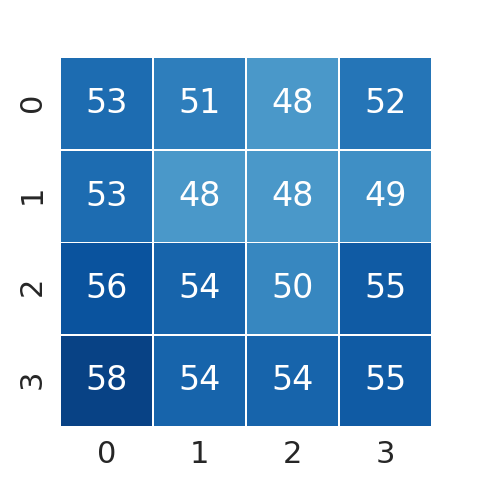}\label{fig:heatmpa-mnist-4x4-075-unif}}\hspace{0.2cm}
\subfloat[\scriptsize{100\% - Uniform }]
         {\includegraphics[width=0.23\textwidth]{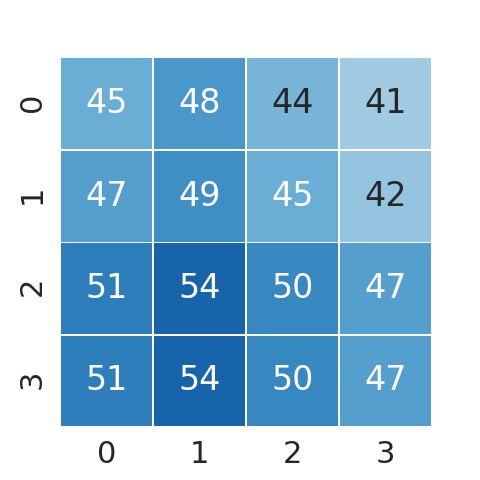}\label{fig:heatmpa-mnist-4x4-100-unif}}

\subfloat[\scriptsize{25\% - ES-(1+1)}]
         {\includegraphics[width=0.23\textwidth]{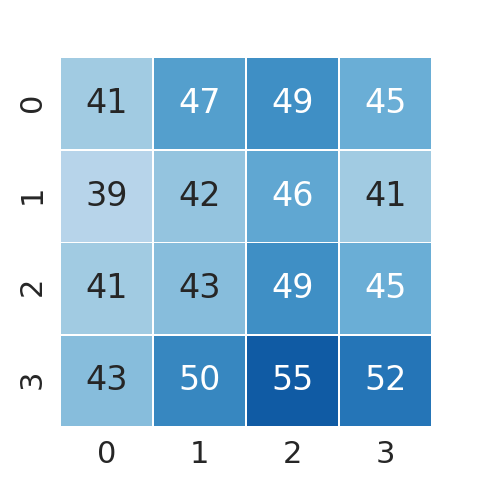}\label{fig:heatmpa-mnist-4x4-025-op}}\hspace{0.2cm}
\subfloat[\scriptsize{50\% - ES-(1+1)}]
         {\includegraphics[width=0.23\textwidth]{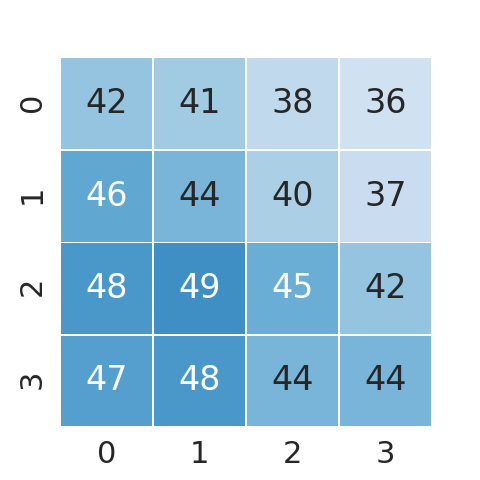}\label{fig:heatmpa-mnist-4x4-050-op}}\hspace{0.2cm}
\subfloat[\scriptsize{75\% - ES-(1+1)}]
         {\includegraphics[width=0.23\textwidth]{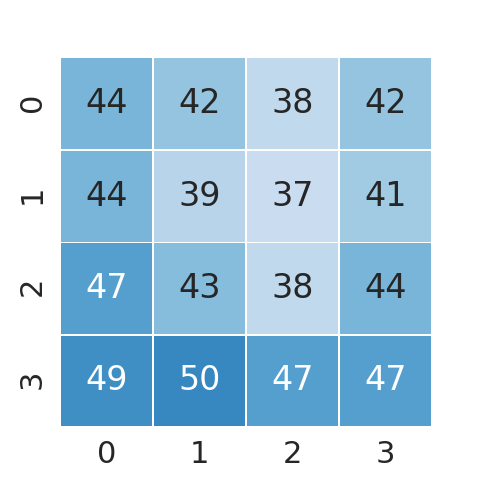}\label{fig:heatmpa-mnist-4x4-075-op}}\hspace{0.2cm}
\subfloat[\scriptsize{100\% - ES-(1+1)}]
         {\includegraphics[width=0.23\textwidth]{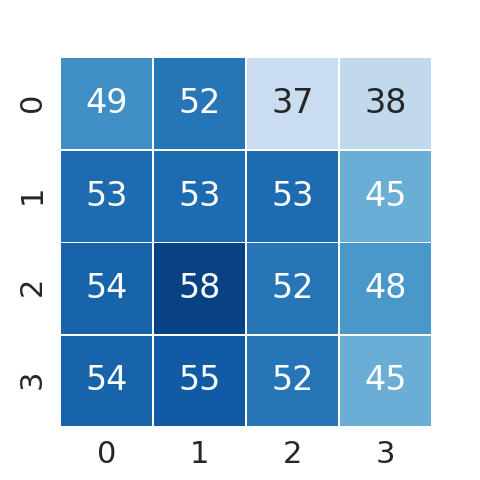}\label{fig:heatmpa-mnist-4x4-100-op}}

        \caption{\small{FID score distribution through a given grid at the end of an independent run for MNIST-\xGrid{4}. Lighter blues represent lower (better) FID scores. The first row (a, b, c, and d) illustrates the FIDs of each ensemble by using uniform mixture weights. The third row (e, f, g, and h) shows the FIDs of each ensemble by using the weights computed by the ES-(1+1).}} 
  \label{fig:heatmaps}
\end{figure}

Finally, the results in Table ~\ref{tbl:best-mean-fids} and Table~\ref{tbl:mean-mean-fids} are less competitive (higher FIDs) than the ones presented in Table~\ref{tbl:best-fids} and Table~\ref{tbl:mean-fids}, respectively. 
This delves on the idea of the improvements on the results when \ensemble{s} are used. 
Fig.~\ref{fig:heatmaps} illustrates the FID scores distribution at the end of an independent run of MNIST-\xGrid{4}.
Note that we cannot compare the results among the different data sizes since we have selected a random independent run, and therefore, the these results do not follow the general observation discussed above. 
The impact of the \ensemble optimization is shown in this figure. 
Here, we can observe how the ES(1+1) optimizes the \ensemble FID values for each cell and manages to improve most of them (the best cell of the grid is always improved). 

\section{Conclusions and Future Work}
\label{sec:conclusions}

The use of less training data reduces the storage requirements,
both disk and RAM while depending on the ensemble of generators to limit possible loss in performance from the reduction of training
data. In the case of \method there is also the potential benefit of the implicit communication that comes from the training on overlapping neighborhoods and updating the cell with the best generator after a training epoch.  Our method, \method, for spatially
distributed evolutionary GAN training makes use of information
exchange between neighborhoods to generate high performing generator
mixtures. The spatially distributed grids allow training with less of
the dataset because of signal propagation leading to exchange of
information and improved performance when training data is reduced
compared to ordinary parallel GAN training. In addition, the ensembles lose performance when the training data is reduced, but they are surprisingly robust with 75\% of the data. 

Future work will investigate the impact of distributing different
modes(e.g. classes) of the data to different cells. In addition, more
data sets will be evaluated, as well as more fine grained reductions
in amount of training data.

\bibliographystyle{spmpsci}
\bibliography{bibliography}

\begin{thebibliography}{10}
\providecommand{\url}[1]{{#1}}
\providecommand{\urlprefix}{URL }
\expandafter\ifx\csname urlstyle\endcsname\relax
  \providecommand{\doi}[1]{DOI~\discretionary{}{}{}#1}\else
  \providecommand{\doi}{DOI~\discretionary{}{}{}\begingroup
  \urlstyle{rm}\Url}\fi

\bibitem{morse2016simple}
Morse~et al., G.: Simple evolutionary optimization can rival stochastic
  gradient descent in neural networks.
\newblock In: GECCO, pp. 477--484. ACM (2016)

\bibitem{schmiedlechner2018towards}
Al-Dujaili, A., Schmiedlechner, T., Hemberg, E., O'Reilly, U.M.: Towards
  distributed coevolutionary {GANs}.
\newblock In: AAAI 2018 Fall Symposium (2018)

\bibitem{Ash2018Reckless}
Al-Dujaili, A., Srikant, S., Hemberg, E., O'Reilly, U.M.: On the application of
  {D}anskin's theorem to derivative-free minimax optimization.
\newblock Int. Workshop on Global Optimization  (2018)

\bibitem{Ambrogioni2019kGANsEO}
Ambrogioni, L., G{\"u}çl{\"u}, U., van Gerven, M.: k-gans: Ensemble of
  generative models with semi-discrete optimal transport.
\newblock ArXiv \textbf{abs/1907.04050} (2019)

\bibitem{arnaldo2015bring}
Arnaldo, I., Veeramachaneni, K., Song, A., O'Reilly, U.M.: Bring your own
  learner: A cloud-based, data-parallel commons for machine learning.
\newblock IEEE Computational Intelligence Magazine \textbf{10}(1), 20--32
  (2015)

\bibitem{arora2017generalization}
Arora, S., Ge, R., Liang, Y., Ma, T., Zhang, Y.: Generalization and equilibrium
  in generative adversarial nets (gans).
\newblock arXiv preprint arXiv:1703.00573  (2017)

\bibitem{arora2017gans}
Arora, S., Risteski, A., Zhang, Y.: Do {GAN}s learn the distribution? some
  theory and empirics.
\newblock In: International Conference on Learning Representations (2018).
\newblock \urlprefix\url{https://openreview.net/forum?id=BJehNfW0-}

\bibitem{Grover2017BoostedGM}
Grover, A., Ermon, S.: Boosted generative models.
\newblock ArXiv \textbf{abs/1702.08484} (2017)

\bibitem{Hardy2018MDGANMG}
Hardy, C., Merrer, E.L., Sericola, B.: Md-gan: Multi-discriminator generative
  adversarial networks for distributed datasets.
\newblock 2019 IEEE International Parallel and Distributed Processing Symposium
  (IPDPS) pp. 866--877 (2018)

\bibitem{Herrmann1999Genetic}
Herrmann, J.W.: A genetic algorithm for minimax optimization problems.
\newblock In: CEC, vol.~2, pp. 1099--1103. IEEE (1999)

\bibitem{Martin2017GANs}
Heusel, M., Ramsauer, H., Unterthiner, T., Nessler, B.: Gans trained by a two
  time-scale update rule converge to a local nash equilibrium.
\newblock arXiv preprint arXiv:1706.08500  (2017)

\bibitem{heusel2017gans}
Heusel, M., Ramsauer, H., Unterthiner, T., Nessler, B., Klambauer, G.,
  Hochreiter, S.: {GAN}s trained by a two time-scale update rule converge to a
  nash equilibrium.
\newblock arXiv preprint arXiv:1706.08500 \textbf{12}(1) (2017)

\bibitem{Hillis1990Coev}
Hillis, W.D.: Co-evolving parasites improve simulated evolution as an
  optimization procedure.
\newblock Physica D: Nonlinear Phenomena \textbf{42}(1), 228 -- 234 (1990).
\newblock \doi{https://doi.org/10.1016/0167-2789(90)90076-2}

\bibitem{hodjat2014maintenance}
Hodjat, B., Hemberg, E., Shahrzad, H., O’Reilly, U.M.: Maintenance of a long
  running distributed genetic programming system for solving problems requiring
  big data.
\newblock In: Genetic Programming Theory and Practice XI, pp. 65--83. Springer
  (2014)

\bibitem{Husbands1994Distributed}
Husbands, P.: Distributed coevolutionary genetic algorithms for multi-criteria
  and multi-constraint optimisation.
\newblock In: AISB Workshop on Evolutionary Computing, pp. 150--165. Springer
  (1994)

\bibitem{lecun1998mnist}
LeCun, Y.: The mnist database of handwritten digits.
\newblock http://yann. lecun. com/exdb/mnist/  (1998)

\bibitem{li2017towards}
Li, J., Madry, A., Peebles, J., Schmidt, L.: Towards understanding the dynamics
  of generative adversarial networks.
\newblock arXiv preprint arXiv:1706.09884  (2017)

\bibitem{liu2015faceattributes}
Liu, Z., Luo, P., Wang, X., Tang, X.: Deep learning face attributes in the
  wild.
\newblock In: Proceedings of International Conference on Computer Vision (ICCV)
  (2015)

\bibitem{loshchilov2013surrogate}
Loshchilov, I.: Surrogate-assisted evolutionary algorithms.
\newblock Ph.D. thesis, University Paris South Paris XI; National Institute for
  Research in Computer Science and Automatic-INRIA (2013)

\bibitem{Mitchell2006Coevolutionary}
Mitchell, M.: Coevolutionary learning with spatially distributed populations.
\newblock Computational Intelligence: Principles and Practice  (2006)

\bibitem{salimans2017evolution}
Salimans, T., Ho, J., Chen, X., Sutskever, I.: Evolution strategies as a
  scalable alternative to reinforcement learning.
\newblock arXiv:1703.03864  (2017)

\bibitem{clune2017uber}
Stanley, K.O., Clune, J.: Welcoming the era of deep neuroevolution - uber
  engineering blog.
\newblock https://eng.uber.com/deep-neuroevolution/ (2017)

\bibitem{Tolstikhin2017AdaGANBG}
Tolstikhin, I.O., Gelly, S., Bousquet, O., Simon-Gabriel, C.J., Sch{\"o}lkopf,
  B.: Adagan: Boosting generative models.
\newblock In: NIPS (2017)

\bibitem{toutouh2019}
Toutouh, J., Hemberg, E., O'Reilly, U.M.: Spatial evolutionary generative
  adversarial networks.
\newblock In: GECCO (2019)

\bibitem{wang2018evolutionary}
Wang, C., Xu, C., Yao, X., Tao, D.: Evolutionary generative adversarial
  networks.
\newblock arXiv preprint arXiv:1803.00657  (2018)

\bibitem{Wang2016EnsemblesOG}
Wang, Y., Zhang, L., van~de Weijer, J.: Ensembles of generative adversarial
  networks.
\newblock ArXiv \textbf{abs/1612.00991} (2016)

\bibitem{wierstra2008natural}
Wierstra, D., Schaul, T., Peters, J., Schmidhuber, J.: Natural evolution
  strategies.
\newblock In: Evolutionary Computation, 2008. CEC 2008.(IEEE World Congress on
  Computational Intelligence). IEEE Congress on, pp. 3381--3387. IEEE (2008)

\bibitem{Williams2005Investigating}
Williams, N., Mitchell, M.: Investigating the success of spatial coevolution.
\newblock In: Proceedings of the 7th annual conference on Genetic and
  evolutionary computation, pp. 523--530. ACM (2005)

\bibitem{Zhong2019RethinkingGC}
Zhong, P., Mo, Y., Xiao, C., Chen, P., Zheng, C.: Rethinking generative
  coverage: A pointwise guaranteed approach.
\newblock ArXiv \textbf{abs/1902.04697} (2019)

\end{thebibliography}

\end{document}